\documentclass[11pt,a4paper]{article}
\usepackage[hyperref]{acl2021}
\usepackage{times}
\usepackage{latexsym}

\usepackage{microtype}

\date{}
\usepackage{times}
\usepackage{latexsym}

\usepackage{microtype}

\usepackage{xspace,mfirstuc,tabulary}
\usepackage{times}
\usepackage{latexsym}

\usepackage{microtype}

\usepackage{url}
\usepackage{booktabs}       
\usepackage{amsfonts}       
\usepackage{nicefrac}       
\usepackage{microtype}      
\usepackage{algorithmic}
\usepackage{algorithm}
\usepackage{wrapfig}
\usepackage{lipsum}
\usepackage{xspace}
\usepackage{graphicx} 
\usepackage{subfigure} 
\usepackage{caption}
\usepackage{paralist}
\usepackage{float}
\usepackage{tabularx}
\usepackage{multirow}
\usepackage{diagbox}
\usepackage{tikz}
\usepackage{booktabs}
\usepackage{array, makecell} 
\usepackage{natbib}
\usepackage{amsmath,amssymb}
\usepackage[inline]{enumitem}
\usepackage{acronym}
\usepackage[T1]{fontenc}
\usepackage[utf8]{inputenc}
\usepackage[figurename=Fig.]{caption}

\newcounter{todocnt}

\acrodef{RL}{Reinforcement Learning}
\acrodef{PL}{Policy Learning}
\acrodef{DRL}{Deep Reinforcement Learning}
\acrodef{IRL}{inverse reinforcement learning}
\acrodef{SERP}{search engine result page}
\acrodef{IR}{Information Retrieval}
\acrodef{MDP}{Markov Decision Process}
\acrodef{MaxEnt-IRL}{Maximum Entropy Inverse Reinforcement Learning}
\acrodef{DM-IRL}{Distance Minimization Inverse Reinforcement Learning}
\acrodef{MMI}{Maximum Mutual Information} 
\acrodef{DNN}{Deep Neural Networks}
\acrodef{RNN}{Recurrent Neural Networks}
\acrodef{MLP}{Multilayer Perceptron}
\acrodef{MLPs}{Multilayer Perceptrons}
\acrodef{GRU}{Gated Recurrent Net}
\acrodef{TDSs}{Task-oriented Dialogue Systems}
\acrodef{TDS}{Task-oriented Dialogue System}
\acrodef{LU}{language understanding}
\acrodef{DM}{Dialogue Management}
\acrodef{NLG}{natural language generation}
\acrodef{DST}{Dialogue State Tracker}
\acrodef{SOTA}{state-of-the-art}
\acrodef{NLU}{Natural Language Understanding}

\acrodef{DQN}{Deep Q-learning}

\aclfinalcopy 

\title{Data-driven Approach to Uncover User Satisfaction Structure\\ in Multi-turn Dialogues}

\title{A Data-driven Approach to Estimate User Satisfaction Level\\ in Multi-turn Dialogues}

\title{DEUS: A Data-driven Approach to Estimate User Satisfaction \\ in Multi-turn Dialogues}

\author{
Ziming Li \textsuperscript{1},
Dookun Park \textsuperscript{2},
Julia Kiseleva \textsuperscript{3},
Young-Bum Kim \textsuperscript{2},
Sungjin Lee \textsuperscript{2}\\
\textsuperscript{1}University of Amsterdam,
\textsuperscript{2}Amazon Alexa AI,
\textsuperscript{3}Microsoft Research\\
z.li@uva.nl,
julia.kiseleva@microsoft.com,
\{dkpark,youngbum,sungjinl@amazon.com\}
}

\date{}

\begin{document}
\maketitle
\begin{abstract}


Digital assistants are experiencing rapid growth due to their ability to assist users with day-to-day tasks where most dialogues are happening multi-turn. However, evaluating multi-turn dialogues remains challenging, especially at scale. We suggest a context-sensitive method to estimate the turn-level satisfaction for dialogue considering various types of user preferences. The costs of interactions between users and dialogue systems are formulated using a budget consumption concept. We assume users have an initial interaction budget for a dialogue formed based on the task complexity and that each turn has a cost. When the task is completed, or the budget has been exhausted, users quit the dialogue. We demonstrate our method's effectiveness by extensive experimentation with a simulated dialogue platform and real multi-turn dialogues. 
\end{abstract}

\section{Introduction}
\label{sec:introduction}
The recent advances in language understanding techniques~\cite{devlin2018bert, LiuRoberta_2019,clark2020electra} stimulate rapid progress in the development of dialogue systems, which naturally utilized by digital assistants~\cite{liu2017iterative, liu2018adversarial, adiwardana2020towards, roller2020recipes}, such as Amazon Alexa, Apple Siri, and Google Now, which can assist with more and more complex tasks.
%
%
Evaluation is a crucial component while developing a digital assistant which reflects the system's quality and provides insights into required further improvements~\cite{li2019acute,kiseleva2016predicting}.
However, it remains challenging to scale since only users themselves can observe or judge their own experiences. The absence of a reliable understanding of the user feedback loop makes it harder for dialogue systems to adjust its response strategies to maximize user satisfaction.  


Originally,~\citet{walker1997paradise} propose to evaluate the performance of dialogue systems by: (1)~maximizing the task success rate and (2)~minimizing the interaction cost. Interaction cost is composed of efficiency measures (e.g., number of interactions) and qualitative measures (e.g., inappropriate utterance ratio). However, these strategies strongly rely on domain knowledge, which limits their adaptation across different dialogue scenarios.  
It is especially challenging to consider all these metrics and combine them into a single evaluation method in multi-turn dialogues; when tasks are getting more complicated, the interaction histories could consist of half-a-dozen turns. Also, it is likely for the dialogue system to take non-optimal actions during this long interaction. Each dialogue turn may contribute differently to the final dialogue status, denoted by success or failure~\cite{kiseleva2016predicting}. For example, a greeting turn, e.g., \emph{``good morning''}, should have a lower importance weight than a turn providing a specific service, e.g., \emph{``setting the alarm at 7 am''}. Hence, it is reasonable to consider turn priorities while judging the dialogue system's performance. Furthermore, we should take into account task complexity in the evaluation process to normalize the performance. It is easier for dialogue systems to make mistakes when the given task is complex. However, users understand that completing this task takes time, and it will not be easy. So users may bear minor mistakes for complex tasks. In contrast, users expect the digital assistant to finish simple tasks such as \emph{``showing the weather information''} seamlessly. Otherwise, they will not trust the system, as it leads to a poor experience~\cite{kiseleva2016understanding}.

This paper proposes a new \textbf{D}ata-driven approach to \textbf{E}stimate \textbf{U}ser \textbf{S}atisfaction (DEUS) in multi-turn dialogue scenarios widely used by digital assistants. Our primary assumption is when users start a new dialogue
with digital assistants, they reserve task-sensitive \emph{patience budget}, and each turn in dialogue has a cost. The budget is conditioned on the task complexity, and when the task is completed, or the budget has been run out, the user will quit the dialogue. Our method is context-sensitive since user satisfaction is estimated at each turn taking dialogue context as input. The DEUS has a build-in view on a dialogue as a whole because each turn will consume some remaining budget which directly affects the final dialogue status. The main advantage of our data-driven setup is allowing getting rid of relying on a massive amount of domain knowledge to handcraft evaluation metrics.

Our work provides the following contributions:
\begin{enumerate}[leftmargin=*,label=\textbf{C\arabic*},nosep]
  \item We propose a new patience-relevant formulation to interpret the multi-turn interactions between users and dialogue systems;
  \item We invent a new Data-driven method for Estimating User Satisfaction (DEUS) in multi-turn dialogues;
  \item We demonstrate the effectiveness of our approach through extensive experimentation with simulated and realistic dialogues.
\end{enumerate}

\section{Related Work}
\label{sec:related_work}
Our paper is relevant to two strands of research.  
First, we discuss research on estimating user satisfaction for interactive systems, including digital assistants.
Second, our work is related to studied approaches to evaluate dialogue systems.

\textbf{User Satisfaction} aims to measure the quality of user experience while interacting with a system.
\citet{kelly2009methods} defines the satisfaction for interactive systems as \emph{`satisfaction can be understood as the fulfillment of a specified desire or goal'}. Also, \citet{Kelly_ictir_2015} provides evidence that the most complex search tasks have a similar characterization of complex tasks describe by~\citet{Campbell_1988} concerning having multiple interdependent parts that needed to be addressed separately. This definition is also suitable for digital assistants since they interact with users to achieve certain goals through multi-turn dialogues.


Estimating user satisfaction with subjective metrics~\citep{yang2012predicting,hone2001subjective} might be difficult in practice because asking users for explicit feedback may annoy them and hurt user experience unexpectedly. Inferring user satisfaction with automatic metrics is a more desirable solution~\cite{kiseleva2016predicting,kiseleva2016understanding} since it can be performed at scale. 
Earlier, \citet{walker1997paradise} propose a general framework for evaluating and comparing the performance of spoken dialog systems, which can compare agents performing different tasks by normalizing task complexity.
These metrics measure user experience by studying the logged interaction histories from different aspects \citep{moller2005parameters, danieli1995metrics, hastie2012metrics}.
Another view on user satisfaction is reflected by user engagement metrics, which show different aspects of user experience. For instance, they can reflect \emph{user loyalty}~\cite{Song_www_2013}, \emph{user activity}~\citep{Lehmann_cikm_2013} or the absence time~\citep{Dupret_wsdm_2013}. \citet{meng2020predicting} proposed a framework for classifying user engagement into four categories: \emph{fulfillment}, \emph{continuation}, \emph{reformulation} and \emph{abandonment}.

\noindent
\par{\textbf{Dialogue Systems}} are generally separated into two types task-oriented and open-domain. Task-oriented systems usually have clear criteria for evaluation such as inappropriate utterance ratio, turn correction ratio, concept accuracy, and task success rate ~\citep{takanobu2019guided,li2016user,Su2018D3Q,li2020guided, zhu-etal-2020-convlab, lee2019convlab}. Open-domain dialogues are generally harder to evaluate due to their nature~\cite{li2019acute, li2018dialogue, sedoc-etal-2019-chateval}. Despite significant efforts to introduce automatic metrics in this space~\cite{reiter2018structured,novikova2017we,lowe2017towards}, the current standard approach to evaluate open-domain dialogues is to employ human evaluations~\cite{zhang2018personalizing,li2019acute},  
because \citet{liu2016not} demonstrated that automatic metrics correlate very poorly with human judgments.

Our work aims to bridge the gap between the two described research areas by proposing a new method to estimate user satisfaction for multi-turn dialogues scenarios employed by digital assistants.

\section{Method: DEUS}
\label{sec:method}

\setlength{\belowdisplayskip}{0pt} \setlength{\belowdisplayshortskip}{0pt}
\setlength{\abovedisplayskip}{0pt} \setlength{\abovedisplayshortskip}{0pt}

\subsection{Assumption: user patience budget}
\label{assumption:patience}
The digital assistant's goal is to efficiently and effectively help users accomplishing their tasks.
We assume that users can estimate the interaction cost before initiating a new task, depending on its complexity. This cost can be interpreted as a \emph{patience budget} that a user is willing to spend on this task. Each turn in the dialogue consumes part of the initial budget. For simple tasks, such as setting the alarm, the patience budget is relatively low, and a user expects the digital assistant to accomplish this type of task seamlessly in fewer turns. In contrast, if the task is complex and requires multi-turn interactions, such as booking a movie ticket, a user is more patient, and the assistant's minor mistakes are still bearable.    
Given the above assumption, a user terminates her interaction in following situations:
\begin{enumerate}[leftmargin=*,nosep]
\item The user's patience budget has been run out, but the task hasn't been accomplished yet.
\item The task is successfully accomplished.
\end{enumerate}

\subsection{Assumption: user preferences}
\label{assumption:user_preference}
It is natural to assume that users have different preferences, which leads to a difference in their behavior in dialogues.
Hence, patience budget and turn-level costs vary, namely:
\begin{enumerate}[leftmargin=*,nosep]
	\item Users perceive the assistant's actions differently, that leads to discrepant evaluation scores for the same response given a similar dialogue context.
  \item Given the same dialogue context, users can have diverse responses, which leads to different dialogues states an assistant takes in the next step.
\end{enumerate}

\subsection{Task formulation}
A multi-turn dialogue is represented as a trajectory $\zeta$ of interactions between a user and a dialogue agent in the format of a sequence of state-action pairs:

\begin{equation}
\zeta =\{ (s_0, a_0, s_1, a_1 \ldots s_t, a_t \ldots ), goal\},
\end{equation}

\noindent
where $s_t$ is the dialogue state, and $a_t$ denotes the agent's action at time step $t$. We use $goal$ to denote the complete user goal. It is a set of slot-value pairs consisting of two parts: (1)~the constraints on different domain slots or booking requirements, e.g., \textit{Restaurant\_Inform\_Food$=$Thai}; (2)~the slot values that show what the user is looking for, e.g., \textit{Restaurant\_Request\_phone$=$?}. We mark a slot as satisfied when the value of this slot has been informed by the user or provided by the agent, depending on the constraint type. If all the slots of the user goal have been satisfied, the task is completed successfully. 
The dialogue state $s_t$ contained the context information from step $0$ to step $t$. The action $a_t$ is from the pre-defined set, which consists of all actions the dialogue system can execute. 

Our task is two-fold to recover: (1)~a function $f(s_t, a_t)$ that estimates the turn-level user satisfaction; and (2)~a function $b(goal)$ that estimates a patience budget for the user goal. The user satisfaction function $f(s_t, a_t)$ takes the dialogue state and action at time $t$ as input and outputs the turn-level cost.
We feed the user's goal, $goal$, to the function $b(goal)$ to obtain the initial user patience budget.
Next, we show how we can jointly estimate $f(s_t,a_t)$ and $b(goal)$.

\subsection{Training objectives}
\label{sec:objectives}
Given a dialogue with $m$ turns, and its corresponding final status, as \emph{failed} or \emph{successful}, denoted as `$-1$' or `$+1$' respectively, we set the training constraints as follows:
%


\begin{equation}
\resizebox{0.8\hsize}{!}{%
$\sum_{t=1}^mf(s_t, a_t) + b(goal) \geq 0 \textit{ if status}=1,$
\label{eq:success_constraint}
}
\end{equation}

\begin{equation}
\resizebox{0.8\hsize}{!}{%
$\sum_{t=1}^mf(s_t, a_t) + b(goal) < 0 \textit{ if status}=-1.$
}
\label{eq:fail_constraint}
\end{equation}
\noindent
Following our assumptions about user patience described in Section~\ref{assumption:patience}:
\begin{itemize}[leftmargin=*,nosep]
    \item Eq.~\ref{eq:success_constraint} corresponds to the successful dialogues where users still have budget left when they quit dialogues; and
    \item Eq.~\ref{eq:fail_constraint} represents the case of failed dialogues. 
\end{itemize}
For both successful and failed dialogues, the user still has patience left at time step $m-1$, and this brings us an additional constraint:

\begin{equation}
\resizebox{0.6\hsize}{!}{%
$\sum_{t=1}^{m-1}f(s_t, a_t) + b(goal) > 0.$
}
\label{eq:one2last_constraint}
\end{equation}
We assume the patience budget is always positive (Eq.~\ref{eq:goal_v}) and the turn-level cost is negative (Eq.~\ref{eq:f_v}): 
\begin{equation}
b(goal) > 0,
\label{eq:goal_v}
\end{equation}
\begin{equation}
f(s_t, a_t) < v_b < 0,
\label{eq:f_v}
\end{equation}
where $v_b$ is a constant that reflects each dialogue turn has an inherent cost regardless of the $a_t$ quality. We define the dialogue-level satisfaction as the remaining budget when the dialogue is terminated. For failed dialogues, the dialogue-level satisfaction is set to zero.\\
Next, we convert the training constraints to Hinge Loss. Eq.~\ref{eq:success_constraint} and Eq.~\ref{eq:fail_constraint} can be merged to formulate our first training loss:
\begin{equation}
\resizebox{0.92\hsize}{!}{%
$loss_1 = \max (0, -\textit{status} * (\sum_{t=1}^mf(s_t, a_t) + b(goal))).$
}
\label{eq:loss_1}
\end{equation}
Eq.~\ref{eq:one2last_constraint} helps to formalize our second loss as:
\begin{equation}
\resizebox{0.92\hsize}{!}{%
$loss_2 = \max (0, -(\sum_{t=1}^{m-1}f(s_t, a_t) + b(goal))).$
}
\label{eq:loss_2}
\end{equation}
%
Eq.~\ref{eq:goal_v} and Eq.~\ref{eq:f_v} are merged to reflect the third loss:
\begin{equation}
\resizebox{0.8\hsize}{!}{%
$loss_3 = \sum_{t=1}^m\max (0, f(s_t, a_t) - v_b)).$
}
\label{eq:loss_3}
\end{equation}


\noindent
The following setups are used as final training loss:
\begin{equation}
\resizebox{0.65\hsize}{!}{%
$loss_{full} = loss_1 + loss_2 + loss_3.$
}
\label{eq:loss_full}
\end{equation}

\begin{equation}
\resizebox{0.55\hsize}{!}{%
$loss_{light} = loss_1 + loss_3.$
}
\label{eq:loss_light}
\end{equation}
$\textit{loss}_\textit{light}$ is used to verify the effectiveness of the additional constraint denoted by Eq.~\ref{eq:one2last_constraint}.

\subsection{Forward-looking training objectives}
\label{sec:forward-looking}
In Section~\ref{assumption:patience}, we regard the interaction cost as a fixed value, which may not conform to realistic human behavior patterns. In fact, users are forward-looking, and they dynamically adjust the cost estimation that they have to investigate in the ongoing interaction. 
This potential cost is affected by the task complexity as we specified in Section~\ref{assumption:patience}. It is also sensitive to the system's performance that the user has observed so far. If the system performs worse than the user expected, she needs to spend more cost on the remaining slots that have not been satisfied. The consequence is that the user will terminate the dialogue earlier to avoid non-necessary loss if the remaining unsatisfied slots' newly estimated cost is higher than the remaining patience. 
We use $goal'$ to denote the remaining unsatisfied slots in the same dialogue.
$c(\textit{goal}')$ estimates the potential cost that the user needs to invest on the remaining slots based on their experience so far. The true value of $c(\textit{goal}')$ is only available during the data collection step, and a function $c(\cdot)$ is used to approximate this value during training.

For forward-looking users, loss functions are:
\begin{equation}
\resizebox{0.85\hsize}{!}{%
$
\begin{split}
\mbox{}\hspace*{-2mm} loss_{1_{forward}} &= \max (0, -\textit{status} * (\sum_{t=1}^mf(s_t, a_t) \\ 
& + b(goal) - c(goal'))).
\end{split}
$
\hspace*{-1mm}\mbox{}
}
\label{eq:loss_1_forward}
\end{equation}

\begin{equation}
\begin{split}
\mbox{}\hspace*{-2mm} loss_{2_{forward}} = &\max (0, - (\sum_{t=1}^{m-1}f(s_t, a_t) \\
& + b(goal) - c(goal'))).
\end{split}
\hspace*{-1mm}\mbox{}
\label{eq:loss_2_forward}
\end{equation}
$\textit{loss}_{3_{forward}}$ is the same as $\textit{loss}_3$. The key idea is that in the same dialogue, the left patience for the remaining unsatisfied slots should be larger than the newly estimated potential cost. For the training objective, $\textit{loss}_{full_{forward}} = \textit{loss}_{1_{forward}} + \textit{loss}_{2_{forward}} + \textit{loss}_{3_{forward}}$ is the final loss function for forward-looking users. \\

\section{Experiments: simulated setup}
We start from a simulated interactive system, where a rule-based user simulator~\citep{li2016user,lee2019convlab} is built to interact with the dialogue system. The benefit is that we can mimic different user behaviors by manipulating the simulator's rules, e.g., when to terminate the dialogue. Each type of user behavior can be represented with one specific satisfaction function.  This makes it possible for us to compare the recovered satisfaction function with the ground truth. 

\subsection{Experimental Pipeline}
\label{sec:pipeline}


The simulated end-to-end pipeline of deploying a dialogue system online and improve the user experience with estimated user preference consists of four steps: (1) offline training, (2) collecting suboptimal interactions, (3) using DEUS to estimate satisfaction, and (4) dialogue agent retraining.\\
\noindent \textbf{Step 1: Offline training} 
While building a dialogue system, the system designer considers the preferences and habits of the target user group, denoted by $\textit{User}_1$, which can be an individual user or a group of users with similar preferences. The assumption is that the preferences of $\textit{User}_1$ are well studied and given by the satisfaction function $f_1(s_t, a_t)$.
In this experiment, $\textit{User}_1$ is a rule-based simulator and its behavior is defined according to the the turn-level satisfaction function, $f_1(s_t, a_t)$. By incorporating $f_1(s_t, a_t)$ into the dialogue agent's training process, we obtain $\textit{Agent}_1$ that can provide reasonable service to $\textit{User}_1$. \\
\noindent \textbf{Step 2: Collecting suboptimal interactions} Given the dialogue agent trained with the existing user, $\textit{User}_1$, we deploy and expose it to a new user, $\textit{User}_2$. We assume $\textit{User}_2$ has a different turn-level satisfaction function, denoted by $f_2(s_t, a_t)$, against $\textit{User}_1$. Since $\textit{Agent}_1$ is trained for $\textit{User}_1$ and it has no information about $\textit{User}_2$, the interactions between $\textit{Agent}_1$ and $\textit{User}_2$ are not going smoothly, and we define them as \emph{suboptimal interactions}. \\
\noindent \textbf{Step 3: Estimating satisfaction} We use our method DEUS to collected suboptimal interactions to recover the turn-level satisfaction  $f_2(s_t,a_t)$ and patience budget $b_2(goal)$ of $\textit{User}_2$.
The recovered function of turn-level satisfaction $f_2(s_t,a_t)$ is used to estimate overall user experience in multi-turn dialogues. Furthermore, we can reoptimize the dialogue agent by incorporating newly uncovered $f_2(s_t,a_t)$ into the training process, leading to a better performing dialogue agent.\\
\noindent \textbf{Step 4: Dialogue agent retraining} Depending on how we train our dialogue agent in Step 1, we have different ways to reoptimize it. For rule-based systems, we can incorporate the newly recovered knowledge about $\textit{User}_2$ into the system by explicitly adjusting the response rules. In our work, the dialogue agent is trained using \ac{RL} paradigm~\citep{sutton2018reinforcement}, and we can easily replace the satisfaction function with $f_2(\cdot)$ during the training process. We expect the reoptimized system, $\textit{Agent}_2$, can achieve better performance than the performance shown in Step 2.

\subsection{Training setup}
\textbf{MultiWOZ} is a multi-domain dialogue dataset spanning $7$ distinct domains: Attraction, Hospital, Police, Hotel, Restaurant, Taxi, Train, and it consists of $10,438$ dialogues~\citep{budzianowski2018multiwoz}. This dataset's interactions mainly mimic how a dialogue agent helps tourists accomplish some tasks, such as booking a restaurant or recommending a hotel with specific requirements. 

\noindent
\textbf{Convlab} is an open-source multi-domain end-to-end dialogue system platform offering the annotated MultiWOZ dataset and associated pre-trained reference models~\citep{lee2019convlab}. A rule-based user simulator is embedded in this platform.
We use this platform to validate the dialogue agent's performance concerning different user groups with diverse preferences and interests. 
The ConvLab action space consists of $300$ predefined action templates, such as \textit{request(date, food\_type)} and \textit{inform(phone, address)} \footnote{\textit{request(date, food\_type)} means the system is asking for the information about the date and the food type that the user prefers, while \textit{inform(phone, address)} denotes that the system is providing the user with the phone number and address.}.

\noindent \textbf{Dialogue agent training} We use \ac{DQN}~\citep{mnih2015human}, which is an off-policy \ac{RL} algorithm, to train the dialogue agent in Step 1 and Step 4.
We implemented the \ac{DQN} algorithm by utilizing the \ac{RL} training modules in ConvLab.

\noindent \textbf{$\textit{User}_1$} is assumed to have no specific preferences about the interactions, and she only cares if her task can be accomplished successfully with fewer turns. Hence, we handcraft the corresponding user satisfaction function $f_1(\cdot)$ accordingly. Assuming the interaction between the agent and $\textit{User}_1$ terminates at time step $m$, $f_1(\cdot)$ is defined as:

\begin{equation}
\resizebox{0.85\hsize}{!}{%
  $f_1(s_t, a_t) =
    \begin{cases}
      \left|r\right| & \text{if } t=m \text{ and }  \textit{ status}=1,\\
      -\left|r\right| & \text{if } t=m \text{ and }  \textit{ status}=-1,\\
      \left|p\right| & \text{if } t < m.
    \end{cases}     $ 
}
\end{equation}
The value of $p$ is much smaller than $r$. At each turn before the terminal state, a small negative value $-\left|p\right|$ is returned to encourage shorter interactions. `$\textit{status}=1$' corresponds to successful dialogues while `$\textit{status}=-1$' means the dialogue is failed.

\noindent
\textbf{$\textit{User}_2$} is assumed to be different from $\textit{User}_1$, as she not only cares if the task can be finished successfully, but also the interaction experience. 
We build a user simulator with the following properties:

\begin{enumerate}[leftmargin=*,label=\textbf{Pr.\arabic*},nosep]
  \item \label{item:pro_1}  $\textit{User}_2$ dislikes dialogue agents that provide or request too much information in the same turn; we use the number of slots, denoted by $n_{slot}(a_t)$, to measure the amount of information contained in a dialogue turn $a_t$. 
  \item \label{item:pro_2} $\textit{User}_2$ still prefers shorter dialogues;
 \item \label{item:pro_3}  $\textit{User}_2$ prefers to answer or ask only one question, corresponding to one slot, at each turn.
\end{enumerate}
We design a satisfaction function $f_2(\cdot)$ to satisfy~\ref{item:pro_1} and~\ref{item:pro_2}:
\begin{equation}
\label{eq:r_2}
f_2(s_t, a_t) = -n_{slot}(a_t) -1.
\end{equation}
For example, the slot number of action \emph{request(movie\_date, number\_of\_person)} equals to $2$. 
The penalizing factor `$-1$' in Eq.~\ref{eq:r_2} encourages shorter dialogues corresponding to~\ref{item:pro_2}. As for~\ref{item:pro_3}, we implemented it by manipulating the responding rules in the user simulator.
\\
\noindent
\textbf{$\textit{User}_3$} is designed to be forward-looking as described in Sec.~\ref{sec:forward-looking}. After each turn, $\textit{User}_3$ will estimate the potential cost to complete remaining unsatisfied slots.
This property is incorporated to the rule-based user simulator to collect dialogues. Except this forward-looking feature, $\textit{User}_3$'s satisfaction function $f_3(\cdot)$ is the same as $f_2(\cdot)$.Since $goal$ and $goal'$ denote the whole set of slots in the user goal and the remaining unsatisfied slots after few turns, we define $goal-goal'$ as slots that have been satisfied already. Assuming $k$ turns have passed, the potential cost $c(goal')$ is defined as:
\begin{equation}
c(goal') = \frac{\sum_{t=1}^kf_3(s_t, a_t)}{b(goal-goal')} * b(goal'),
\label{eq:forward_function}
\end{equation}
where $\sum_{t=1}^kf_3(s_t, a_t)$ refers to the budget that user has already spent on the satisfied slots. If the system performs worse than the user expects, this will lead to a higher value of $\frac{\sum_{t=1}^kf_3(s_t, a_t)}{b(goal-goal')}$. In this  case, the user will quit the interaction earlier because of high estimated potential cost.
\\
\noindent
\textbf{Initial patience budget} is defined as the sum of slot number $n_{slot}$ and domain number $n_{domain}$\footnote{The number of involved domains in the user goal, $goal$, is given by domain number $n_{domain}(goal)$. There are $7$ domains in the MultiWOZ dataset.} in the given goal, which can be denoted as:
\begin{equation}
\begin{split}
\mbox{}\hspace*{-2mm} b(\textit{goal}) = n_{slot}(\textit{goal})  +  n_{domain}(\textit{goal}).
\end{split}
\hspace*{-2mm}\mbox{}
\end{equation}
We assume $\textit{User}_1$, $\textit{User}_2$ and $\textit{User}_3$  have the same patience budget and it is easy to personalize it by incorporating user profiles.\footnote{We leave this extension as future work.}  We adopt \ac{MLPs} to approximate the turn-level satisfaction, the budget function and the potential cost function for each user separately. 
In terms of the inherent turn cost $v_b$ in Eq.~\ref{eq:f_v}, our hyper-parameter list is $[-0.5, -1.0, -2.0, -10.0]$.

\section{Experiments: human-machine setup}
To show DUES performance on real dialogues, we apply our method to a machine-human dialogues with the provided dialogue-level satisfaction.\\
\noindent
\textbf{DSTC-8} At the 8th Dialog System Technology Challenge~\citep{kim2019eighth}, one of the tasks aims to build end-to-end completion dialog systems based on ConvLab and MultiWoZ dataset.  The submitted systems are evaluated using human-in-the-loop employed through MTurk\footnote{Amazon Mechanical Turk: \url{https://www.mturk.com/}}. After communicating with the simulated dialogue systems, crowd-workers: (1)~provide a 5-scale rating to report on the whole experience, from language understanding and response appropriateness, and (2) judged the status of each dialogue as \emph{failed} or \emph{successful}.
We collected all the completed dialogues between submitted dialogue systems and crowd-workers to build a realistic human-machine dialogue dataset that is suitable for our task. 
The resulted dataset consists of text-based dialogues and their corresponding user goals.
There is no turn-level user satisfaction in this dataset but we know the dialogue-level scores. We check the correlation between the remaining budget when the dialogue is terminated and the 5-scale dialogue-level satisfaction to validate if our method can provide a reliable judging service. 
The resulted dataset consists of 1200 dialogues in the training set, 150 in the validating set, and 150 in the testing set. 

\noindent
\textbf{Dialogue representations} Since the DSTC-8 data is text-based, there is no available structured representation for the dialogue states and actions. We fine-tune a BERT~\citep{devlin2018bert} model to obtain the representations for state-action pairs. The dataset for fine-tuning is built by rewriting the collected dialogues between a well-trained dialogue agent and a user simulator using the \ac{NLG} module in ConvLab.
We train the BERT base model to predict if the given utterance is the dialogue history's subsequent turn.

\section{Results: a simulated setup}
\subsection{Success rate of different dialogue agents}
According Step 1 in Sec.~\ref{sec:pipeline}, we train a dialogue agent $\textit{Agent}_1$ for the existing user -- $\emph{User}_1$. Since her satisfaction function $f_1(\cdot)$ is known, we can achieve success rate as high as $87.9\%$ (Tab.~\ref{Tab.:success_rate}). 

As for Step 2, we pair $\textit{Agent}_1$ with an unseen user $\emph{User}_2$, and the success rate drops to $11.1\%$ (Tab.~\ref{Tab.:success_rate}). One explanation of such low performance is that $\emph{User}_2$ has specific preferences expressed by yet unknown $f_2 (\cdot)$ and inappropriate agent's actions exhaust her patience budget quickly. Another reason is $\emph{User}_2$ asks or answers only one question at each turn, which leads to longer dialogues and is more likely to exceed the maximum allowed number of turns. These two reasons lead to unsuccessful dialogues between  $\emph{User}_2$ and $\textit{Agent}_1$. We use collected dialogues to estimate $f_2(\cdot)$. According to Eq.~\ref{eq:r_2}, the $f_2(\cdot)$ is discrete function. We report the frequency of different satisfaction values in the collected dialogues in Tab.~\ref{Tab.:vb_0-5_l3}. 
We pair $\textit{User}_3$ who is forward-looking (Sec.~\ref{sec:forward-looking}) with $\textit{Agent}_1$ and collect a bunch of interaction trajectories. Because $\textit{User}_3$ is forward-looking, $\textit{Agent}_1$ achieves even lower success rate --  $7.6\%$, as shown in Tab.~\ref{Tab.:success_rate}. 

\begin{table}[t]
  \centering
  \resizebox{0.95\columnwidth}{!}{

 \begin{tabular}{l|cccc}
\hline
\diagbox{\textbf{User}}{\makecell{\textbf{Success rate}}}{\makecell{\textbf{Agent}}} & \makecell{\textbf{Agent 1}} & \makecell{\textbf{Agent 2}} & \makecell{\textbf{Agent 3}} & \makecell{\textbf{Agent 4}} \\
\hline
\makecell{\textbf{User 1}} & 87.9\% & - & - & -\\
\hline
\makecell{\textbf{User 2}} & 11.1\% & 40.5\% & - &- \\
\hline
\makecell{\textbf{User 3}} & 7.6\% & - & 31.8\% & 26.7\%\\
\hline
 \end{tabular}
  }
\caption{The success rate of different dialogue agents with three users. Agent 1 is the dialogue agent trained with User 1; Agent 2 is the retrained agent according to the recovered satisfaction function of User 2; Agent 3 and Agent 4 are retrained agents according to the recovered satisfaction function of User 3.}
\label{Tab.:success_rate}
\vspace{-0.25in}   
\end{table}


\begin{figure}[ht]
\centering
   \includegraphics[clip, width=0.95\columnwidth]{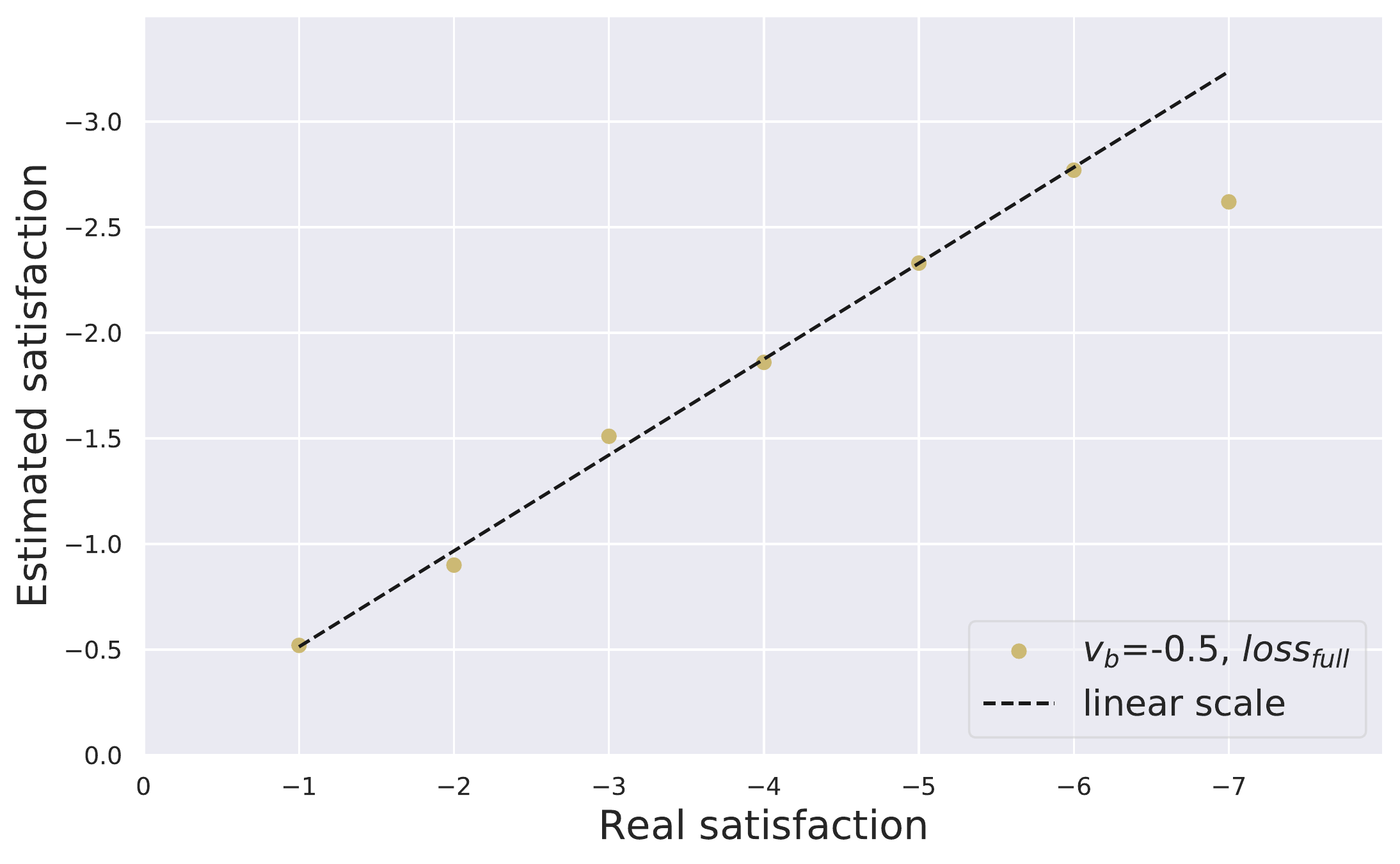}
   \caption{The estimated satisfaction with $v_b = -0.5$ and $\textit{loss}_\textit{full}.$}
   \label{fig:vb_0-5_l3}
\vspace{-0.25in}   
\end{figure} 

\begin{figure}[ht]
\centering
   \includegraphics[clip, width=0.95\columnwidth]{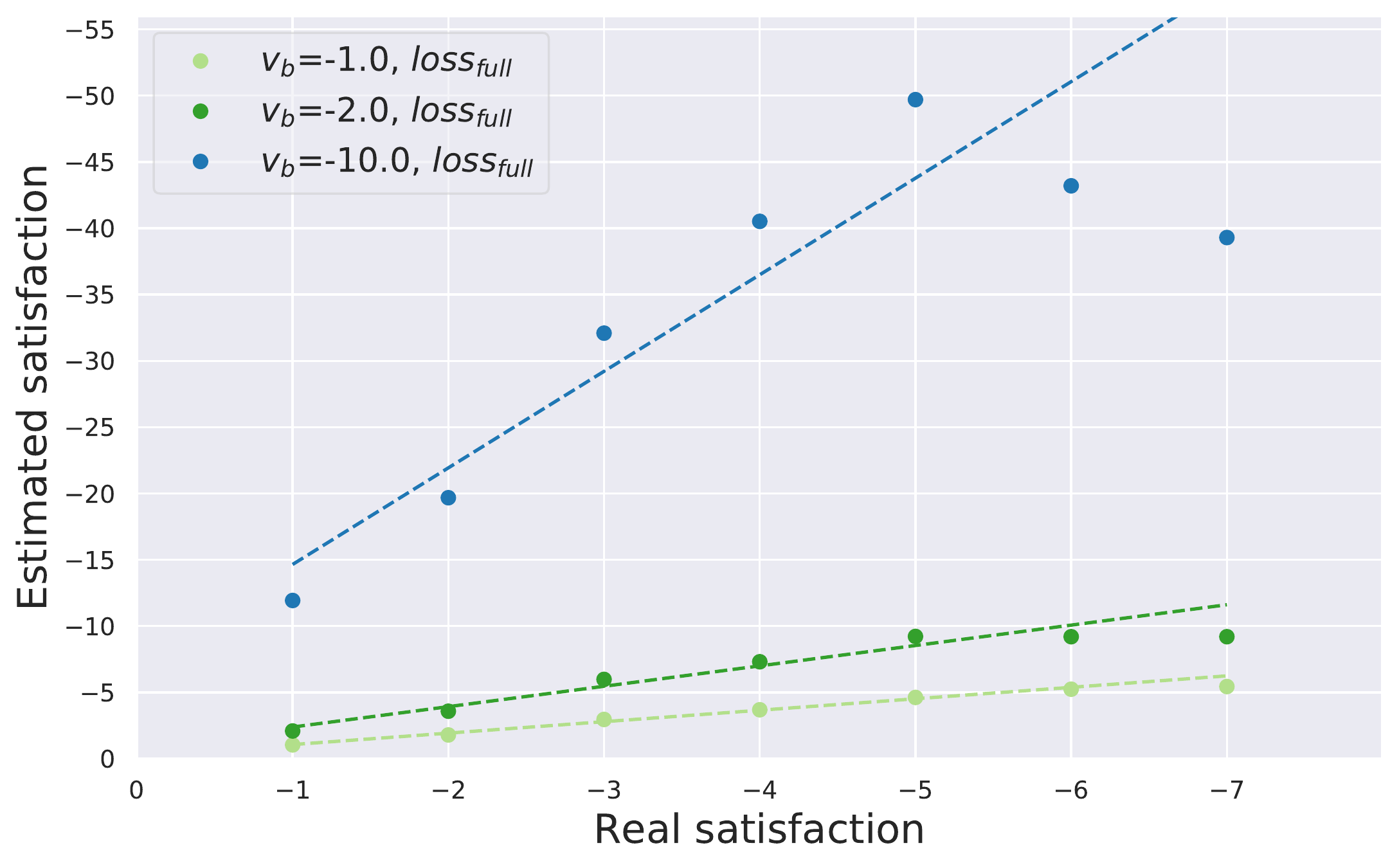}
   \caption{The estimated satisfaction with different inherent turn cost. The loss function is $\textit{loss}_\textit{full}$. Dash lines denote line scale relations.  }
   \label{fig:all_in_one_full}
\vspace{-0.15in}   
\end{figure}


\noindent
\subsubsection{Estimating user satisfaction}
\noindent
\textbf{Estimating user satisfaction various inherent turn costs} 
Tab.~\ref{Tab.:vb_0-5_l3} presents estimated satisfaction values when the inherent cost $v_b$ is $-0.5$ and the training loss is $\textit{loss}_\textit{full}$ defined in Eq.~\ref{eq:loss_full}. To analyze further the correlation between the estimated satisfaction and true satisfaction, we plot them in Fig.~\ref{fig:vb_0-5_l3}. The real satisfaction values can be obtained by linearly rescaling the estimated values. The estimated value for real satisfaction $-7.0$ is an outlier because we only have $2\%$ of dialogue turns that have satisfaction $-7.0$ in the training data, as shown in Tab.~\ref{Tab.:vb_0-5_l3}.
To verify if the detected linear correlation between estimated satisfaction and true one holds for other values of $v_b$, Fig.~\ref{fig:all_in_one_full} plots the corresponding correlations when $v_b$ equals $-1.0$, $-2.0$, or $-10.0$. We observe that the linear relation between estimated and true satisfaction values still exists for the three chosen $v_b$. Interestingly we precisely recover the ground truth satisfaction when the inherent cost is set to $-1.0$. This is because for $\textit{User}_2$, we add an inherent cost, $-1.0$, for each dialogue turn to encourage shorter dialogues, as shown in Eq.~\ref{eq:r_2}. Since we have no access to the true inherent turn cost for realistic dialogues, we can regard it as a hyper-parameter.

\begin{table}
  \centering
  \resizebox{0.95\columnwidth}{!}{
\begin{tabular}{l*{8}{c}}
\toprule
\textbf{Real satisfaction} &  -1.0   &  -2.0  &  -3.0  &  -4.0   &  -5.0  &  -6.0 & -7.0\\
\midrule     
\textbf{Freq in data(\%)}   &33.25 &37.00 &7.76   &14.72 &7.11 &0.14   &0.02 \\
\midrule
\textbf{Estimated sat.} &  -0.52   &  -0.90  &  -1.51  &  -1.86   &  -2.33  &  -2.77 & -2.62\\
\midrule          
\textbf{Std}   &0.07 &0.08 &0.55   &0.2 &0.27 &0.43   &0.15 \\
\bottomrule
\end{tabular} 
}
\caption{The freq. of different satisfaction values in the data collected as a result of Step 2 and the estimated satisfaction on testing dataset with $v_b = -0.5$ and $\textit{loss}_\textit{full}$. } 
\label{Tab.:vb_0-5_l3}
\vspace{-0.15in}
\end{table}




\begin{table}[ht]
\vspace{-0.1in}  
  \centering
  \resizebox{1\columnwidth}{!}{
\begin{tabular}{l*{8}{c}}
\toprule
\multicolumn{8}{c}{(a) $\textit{loss}_\textit{full}$, $v_b=-1.0$}\\ 
\midrule
\textbf{Real satisfaction} &  -1.0   &  -2.0  &  -3.0  &  -4.0   &  -5.0  &  -6.0 & -7.0\\
\midrule     
\textbf{Estimated sat.} &  -1.04   &  -1.79  &  -2.96  &  -3.69   &  -4.62  &  -5.24 & -5.44\\
\midrule          
\textbf{Std}   &0.15 &0.16  &0.9   &0.36  &0.44 &0.95   &0.33 \\
\bottomrule
\end{tabular} 
}

\resizebox{1\columnwidth}{!}{
\begin{tabular}{l*{8}{c}}
\toprule
\multicolumn{8}{c}{(b) $\textit{loss}_\textit{light}$, $v_b=-1.0$}\\ 
\midrule
\textbf{Real satisfaction} &  -1.0   &  -2.0  &  -3.0  &  -4.0   &  -5.0  &  -6.0 & -7.0\\
\midrule     
\textbf{Estimated sat.} &  -1.41   &  -2.25  &  -4.56  &  -5.14   &  -6.58  &  -7.02 & -3.39\\
\midrule          
\textbf{Std}   &0.46 &0.54 &2.3   &0.94 &0.86 &3.03   &0.13 \\
\bottomrule
\end{tabular} 
}
\caption{The estimated satisfaction on testing dataset with different training losses. The top one is trained with $\textit{loss}_\textit{full}$ while the bottom one is with $\textit{loss}_\textit{light}$.} 
\label{Tab.:vb_1_l2_l3}
\vspace{-0.15in}  
\end{table}

\begin{figure}[h]
\centering
  \includegraphics[clip, width=0.95\columnwidth]{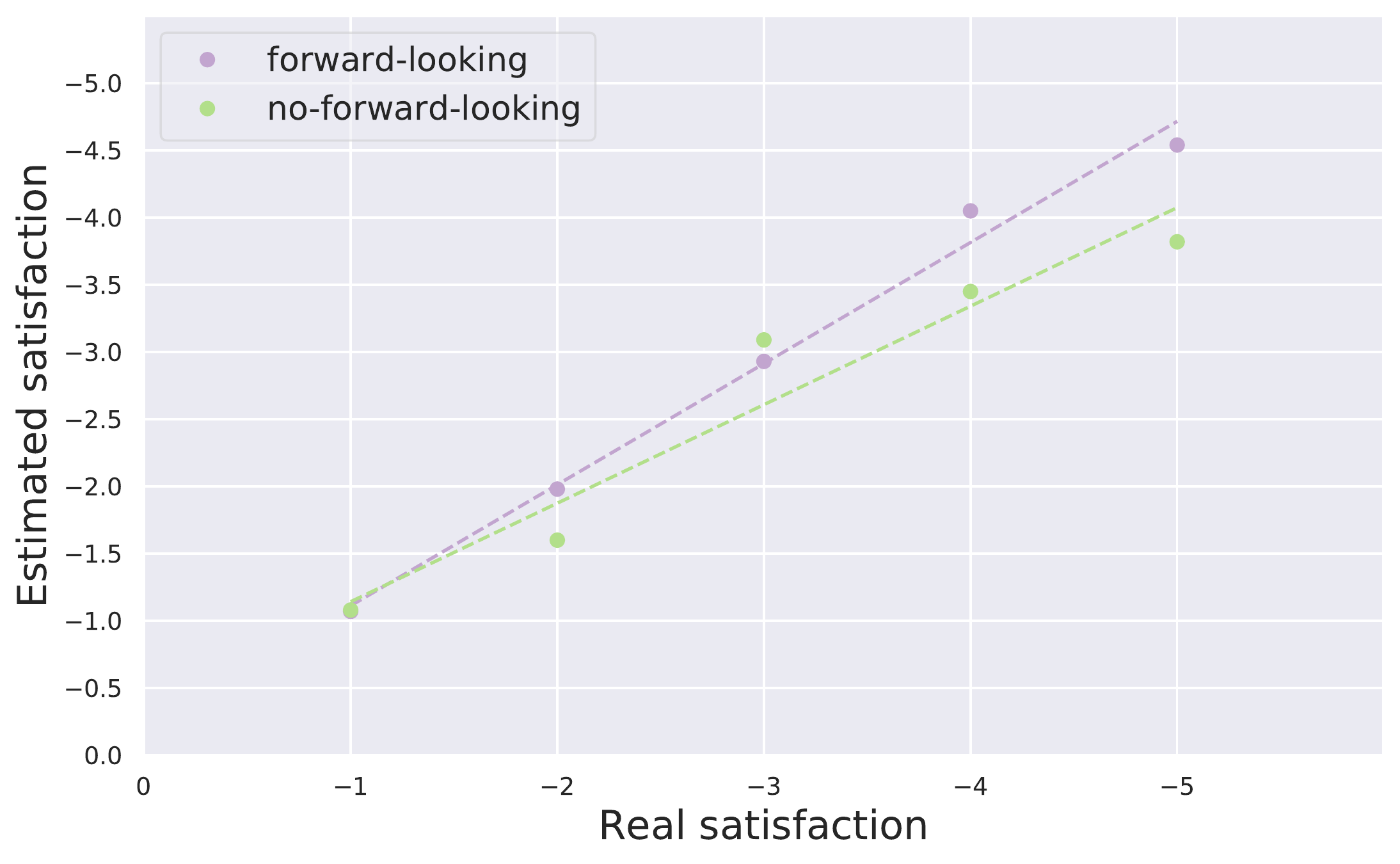}
   \caption{The estimated satisfaction with and without forward-looking functions.}
   \label{fig:vb_1_fl_nofl}
\vspace{-0.15in}     
\end{figure}



\begin{table}[h]
  \centering
  \resizebox{1.\columnwidth}{!}{
\begin{tabular}{l*{6}{c}}
\toprule
\textbf{Setup}   &  $v_b=-1.0$ & $v_b=-2.0$  &  $v_b=-10.0$   & \makecell{forward-\\looking}  &  \makecell{no-forward-\\looking}\\
\midrule     
\textbf{Acc} &  0.982  &  0.974  &  0.975   &  0.872  &  0.830\\
\bottomrule
\end{tabular} 
}
\caption{The accuracy of predicting if the user will terminate the interaction following the patience assumption: for successful dialogues, the remaining budget is non-negative while negative for failed dialogues. The loss function is $\textit{loss}_\textit{full}$.} 
\label{Tab.:acc}
\vspace{-0.25in}  
\end{table}

\begin{figure*}[t]
\centering
   \includegraphics[clip, width=1\linewidth]{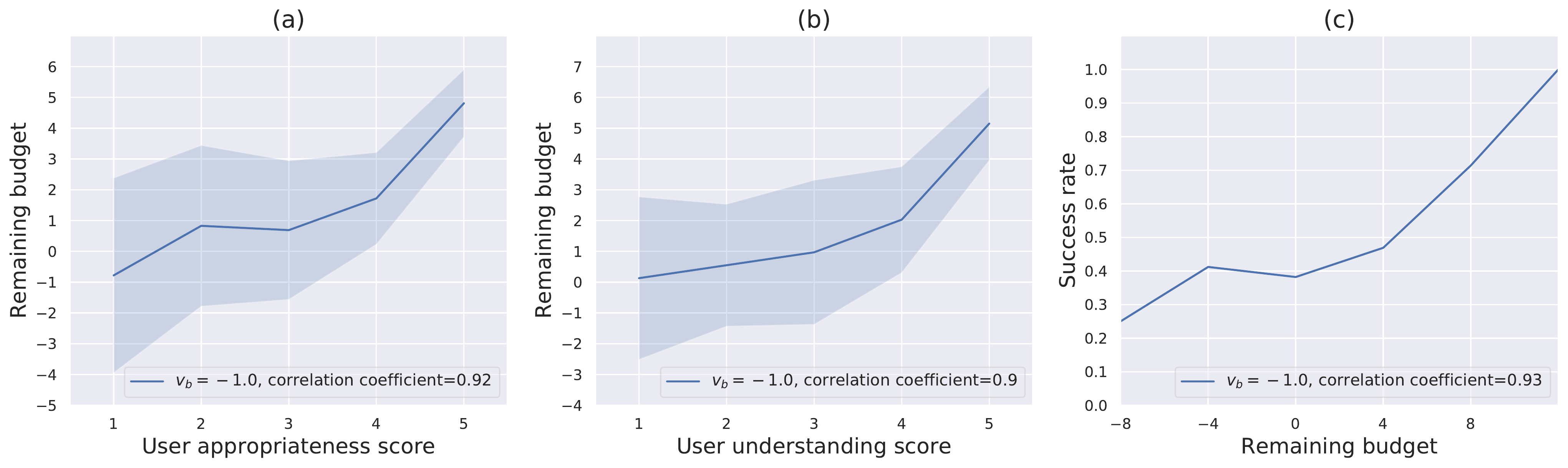}
   \caption{The results on DSTC8 dataset. (a): the correlation between remaining budget and appropriateness score; (b): the correlation between remaining budget and understanding score; (c): the correlation between remaining budget and success rate. Confidence interval is not applicable in (c) because the success rate is at dataset-level.}
   \label{fig:budget_dstc8}
\vspace{-0.15in}     
\end{figure*} 

\noindent
\textbf{Estimating user satisfaction with various losses} 
We report the results with $\textit{loss}_\textit{light}$ (Eq.~\ref{eq:loss_full}) and $\textit{loss}_\textit{full}$ (Eq.~\ref{eq:loss_light}) in Tab.~\ref{Tab.:vb_1_l2_l3} when the inherent turn cost is $-1.0$.\footnote{We choose $v_b=-1$ in this comparison because the true value of $v_b$ for $\textit{User}_2$ is exactly $-1.0$. In this way, we can skip the line rescaling step before comparisons.} It is obvious that the quality of estimated satisfaction has dropped and the standard deviation is getting higher when the training loss is switched to $\textit{loss}_\textit{light}$. For some satisfaction values ,e.g., $-1.0$ and $-3.0$, we can observe a relatively huge gap between the estimated value and the ground truth when the loss function is $\textit{loss}_\textit{light}$. This finding validates the effectiveness of $loss_2$ and the correctness of the constraint from  Eq.~\ref{eq:one2last_constraint}.

\noindent\textbf{Estimating initial patience budget} According to $\textit{loss}_\textit{full}$, the value of initial user patience budget is tied to the scale of the inherent turn cost $v_b$. This makes it impossible to compare the estimated budget with the ground truth directly. To validate the quality of patience budget function $\textit{b(goal)}$, we check if the dialogue status predicted with $\textit{b(goal)}$ and $f_2(s_t, a_t)$, following Eq.~\ref{eq:success_constraint} and Eq.~\ref{eq:fail_constraint}, is consistent with the real status given by the users. As shown in Tab.~\ref{Tab.:acc}, we can achieve quite high accuracy regardless of the inherent turn cost $v_b$.

\noindent
\textbf{Retraining dialogue agent for $User_2$}
We retrain $\textit{Agent}_1$ by introducing the estimated $User_2$ satisfaction function $f_2(\cdot)$. We denote the newly trained agent as $\textit{Agent}_2$ and report its performance in Tab.~\ref{Tab.:success_rate}. Using estimated satisfaction function improves the success rate up to $40.5\%$ within the same evaluation setup. This proves the possibility of successfully reoptimizing dialogue systems using recovered user satisfaction functions.

\noindent
\textbf{Estimating satisfaction for $User_3$} Given the collected interaction trajectories, we recover the satisfaction functions for $User_3$ with the redesigned loss function $\textit{loss}_{full_{forward}}$. We plot the correlation between estimated satisfaction function and the ground truth of $User_3$ in Fig.~\ref{fig:vb_1_fl_nofl}. Obviously, the linear correlation between estimated satisfaction values and the ground truth still exists for $User_3$ after we incorporate the forward-looking property. Furthermore, we conduct an additional experiment with the same collected data where users are forward-looking. We train the $User_3$ satisfaction function with the original loss function $\textit{loss}_{full}$. Since $\textit{loss}_{full}$ does not have a forward-looking assumption, the linear relation still holds for this newly approximated function, but it is not as accurate as the satisfaction function trained with the forward-looking feature as shown in Fig.~\ref{fig:vb_1_fl_nofl}. We retrain the dialogue systems with these two newly approximated satisfaction functions and denote them as $Agent_3$ and $Agent_4$ respectively. Due to the quality difference of estimated satisfaction function, $Agent_3$ achieves a higher success rate -- $31.8\%$, compared to the performance of $Agent_4$ -- $26.7\%$ (Tab.~\ref{Tab.:success_rate}).
We report the results of estimated patience budget in Tab.~\ref{Tab.:acc}. As expected, the setup that considers forward-looking features  achieves higher accuracy. 

\section{Results: a human-machine setup}
For the DSTC8 data, we have 5-scale ratings for the appropriateness score and understanding score from real users. Since we have no access to the turn-level satisfaction, we compare the remaining budget, calculated as $\sum_{t=1}^mf(s_t, a_t)+b(goal)$, with the dialogue-level satisfaction score to validate if our approach works. As shown in Fig.~\ref{fig:budget_dstc8}(a), the remaining budget increases with the user appropriateness score monotonically. Though the increment is not perfectly linear, the correlation is still relatively strong, given the coefficient value, $0.92$. We can find a similar correlation between the user understanding score and the remaining budget in Fig.~\ref{fig:budget_dstc8}(b). In terms of the relation between the success rate and the remaining budget in Fig.~\ref{fig:budget_dstc8}(c), it is reasonable that successful dialogues are more likely to achieve a higher remaining budget. One unexpected phenomenon is that a percentage of successful dialogues has no budget left when the dialogue ends. The potential reasons are threefold:
\begin{enumerate}[leftmargin=*,nosep]
	\item Crowd-workers are paid to interact with the dialogue agents, and they may not follow the budget assumption;
	\item There are multiple dialogue agents, and it is not guaranteed that they perform consistently;
	\item The DSTC8 dataset is not big enough; we only have $1200$ dialogues for satisfaction estimation.
\end{enumerate}

\section{Conclusion}
\label{sec:conclusion}
This paper developed a novel Data-driven method to Estimate User Satisfaction for multi-turn dialogues (DEUS) commonly happening while users interact with digital assistants. 
We modeled the interactions between users and dialogue systems using a budget consumption setup.
DEUS is context-sensitive since user satisfaction is estimated at each turn, taking dialogue history as input; it has a build-in view on a whole dialogue because each turn consumes some remaining budget affecting the final dialogue status
DEUS can reliably infer user satisfaction given interaction histories. We validated its effectiveness based on a simulated dialogue platform and a real-world multi-turn dialogue. 
Incorporating the user profile information into the learning algorithm to personalize the estimated user satisfaction is a promising direction for future work.

\bibliographystyle{acl_natbib}
\bibliography{naaclhlt2019}

\end{document}